# Role-playing Prompt Framework: Generation and Evaluation

Prompt Framework for Role-playing

Xun Liu

Zhejiang Gongshang University, Sussex Artificial Intelligence Institute, nicotieee@gmail.com

Zhengwei Ni

Zhejiang Gongshang University, School of Information and Electronic Engineering, zhengwei.ni@zjgsu.edu.cn

Large language models (LLMs) exhibit impressive proficiency in natural language generation, understanding user instructions, and emulating human-like language use, which has led to significant interest in their application to role-playing scenarios. However, the manual collection of role-specific script data and the evaluation of model performance are resource-intensive processes. This paper introduces a prompt-based framework designed to leverage GPT's capabilities for the generation of role-playing dialogue datasets and the evaluation of role-playing performance. To validate the effectiveness of the GPT-based generation and evaluation, we further incorporate the recall-oriented Rouge-L metric, providing an additional quantitative measure of performance.

# 1 INTRODUCTION

Large language models (LLMs), such as ChatGPT [1], have demonstrated remarkable capabilities in generating coherent and contextually appropriate natural language, made possible by their pretraining on extensive text corpora. These language models have garnered significant attention due to their impressive performance and wide range of applications. As an innovative application area in entertainment and education [2, 3], role-playing with LLMs aims to minimize the disparity between human language abilities and LLM-based dialogue agents by enabling LLMs to emulate the speaking styles, signature phrases, and characteristics of human or fictional characters. Consequently, this role-playing functionality offers users a more tailored and immersive interaction experience [3, 4].

However, since current open-source LLMs are predominantly pre-trained on general-purpose corpora and lack task-specific data for applications such as role-playing dialogue, they are more likely to underperform in this specific application area in comparison to the most state-of-the-art (SOTA) LLMs, like GPT-4 [4, 5]. To address this limitation and enhance pre-trained LLMs' performance across specialized tasks—such as role-playing—fine-tuning has emerged as an active approach in natural language processing (NLP). Fine-tuning enables models to adapt to previously unseen tasks while minimizing the associated computational and resource costs [6, 7]. In the context of role-playing applications, fine-tuning with adequately curated role-playing dialogue datasets and character profiles allows LLMs to acquire character-specific knowledge and emulate a role's distinctive traits, including catchphrases and behavioral characteristics, which significantly enhances their capability to execute customized role-playing tasks effectively [8]. Existing methods for collecting role-playing dialogue data primarily involve extracting role-specific scripts and constructing character profiles. However, prior research indicates that manually collecting and constructing each role's script can be costly [9], and the performance of fine-tuning the model directly on script data is relatively poor [4].

To address these challenges, several methods have been proposed. ChatHaruhi [8] introduces an approach that enables LLMs to generate dialogue for a target role by utilizing questions from the original stories. Character-LLM [9] constructs dialog data by abstracting each character's profile from relevant sources (like Wiki) and then generating scenes based on the profiles. ROLELLM [4] explores the Context-Instruction method, which leverages state-of-the-art (SOTA) LLMs to generate dialogue from the role's profile and structured historical conversation data.

In this paper, we propose a prompt-based role-playing framework that utilizes GPT-4o [1] for both role-playing dialog data generation and performance evaluation. The framework comprises two main stages, summarized as follows: (1) Data Generation: Building on the demonstrated capabilities of GPT-4 in role-playing tasks [4, 5], we utilize GPT-4o to generate role-playing datasets. Inspired by Khot et al. [10], the role-playing dialog generation task is decomposed into three sub-tasks, each carried out using distinct prompts with GPT-4o. First, GPT-4o is prompted to construct scene plots for each target character. Based on these plots, further prompts are designed to guide GPT-4o to generate role-playing question-and-answer dialogs, which are then built into datasets for subsequent fine-tuning experiments. (2) Performance Evaluation: State-of-the-art large language models, such as GPT-4, have also demonstrated advanced capabilities in achieving evaluation tasks effectively [11]. In this paper, we utilize a GPT-based evaluator to assess the quality of the candidate LLM's generation in role-playing tasks. To enhance flexibility and enable GPT-4o to outperform in this complex task [10], the evaluation task is similarly decomposed into three sub-tasks, each of them is conducted by prompting GPT-4o individually to assess one specific dimension of role-playing performance: Characteristics, Task Response, and Generation Quality.

In the experiments, we employ this framework to construct role-playing dialog datasets and evaluate their effectiveness. We first fine-tune several open-source LLMs on the generated datasets, then prompt GPT-4o to rank the performance of the fine-tuned LLMs. Also, we utilize Rouge-L based metrics [12] to measure the overlap of both original and fine-tuned LLMs' predictions on the benchmark constructed by GPT-4o. Additionally, through our framework, we investigate the effects of different data generation prompts, with each prompt designed using different prompt engineering techniques [13, 14], like in-context learning [15, 16] and chain-of-thought reasoning [17].

## 2 RELATED WORK

### 2.1 Role-playing with LLMs

Recent advancements in LLMs have facilitated their adoption in simulating human behaviours in various applications [9]. In terms of human language skills imitation, role-playing with LLMs has emerged as an innovative and potential approach. This practice typically involves users interacting with LLMs in virtual environments by prompting specific characters [3]. This mode of interaction is not only prevalent in gaming and entertainment [18], but also demonstrates potential in educational, training, and simulation scenarios [2, 9]. However, many open-source LLMs' role-playing capabilities lag behind SOTA close-source LLMs. Furthermore, prior research suggests that the current approach to enhancing the role-playing performance of open-source LLMs, primarily fine-tuning the LLM directly on raw script data in the form of multiple dialogue rounds, yields relatively limited improvements [4], which indicates that the unprocessed character script data are unsuitable for fine-tuning LLMs, it is necessary to further synthesize role-playing dialogue data [4]. Therefore, in this paper, we employ a prompt-based framework to leverage the advanced role-playing capabilities of GPT-4o for the generation of role-playing dialog data. These constructed datasets can be further used to improve the capabilities of open-source LLMs in role-playing tasks. Additionally, we design step-by-step evaluation prompts that enable GPT-4o to assess role-playing performance across multiple dimensions.

### 2.2 Prompt Engineering

A prompt refers to a set of instructions provided by a user or system to a LLM, enabling it to perform customizable interactions and advanced functionalities [14]. Specifically, a prompt defines the task's description, scope, and context, while also providing preliminary rules to guide the subsequent responses of the LLM [11, 14]. According to Jules et al. [14], prompt engineering is an approach to program the LLMs via designed prompts. To enhance GPT-4o's capabilities in role-playing tasks, we design various prompts that are engineered to program GPT-4o to accomplish data generation and evaluation tasks. These prompts were constructed using diverse prompt engineering techniques. In the appendices A, we list all the prompts used in this paper.

## 3 APPROACH

This section outlines our methodology for leveraging a prompt-based framework to enhance the performance of LLMs in role-playing tasks through fine-tuning. First, we prompt GPT-4o to generate role-playing dialogue datasets in a step-by-step manner, including the construction of plots, questions, and answers. Using these datasets, we fine-tune several open-source LLMs, Baichuan2 [19], LLaMA2 [20] and ChatGLM2 [21]. Finally, to evaluate the models' performance improvement in role-playing tasks, we utilize a GPT-based evaluator to assess the fine-tuned LLMs through various dimensions. This assessment validates the effectiveness of the proposed framework in enhancing LLMs' role-playing capabilities.

## 3.1 Data Generation

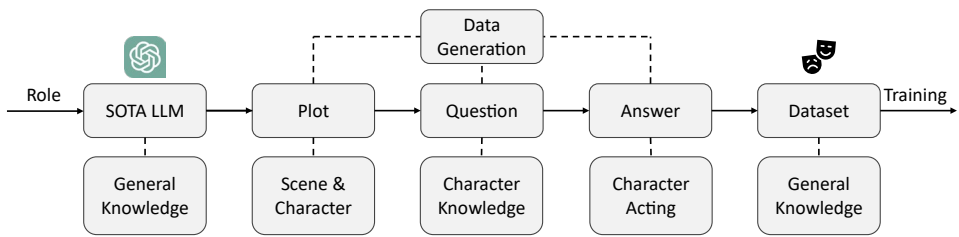

Figure 1: Data-Generation Framework

We aim to leverage the capabilities of SOTA close-source LLMs, such as GPT-4, to generate role-playing dialog data, which can then be constructed into datasets and utilized to enhance the role-playing performance of open-source LLMs. In this paper, as shown in Figure 1, our data-generation framework consists prompting GPT-4o to generate role-playing dialog datasets in a step-by-step manner, including (1) Plot Construction; (2) Question Generation; (3) Answer Generation.

Prior research has demonstrated the remarkable ability of LLMs to perform in-context learning and few-shot prompting [3, 15]. By providing a few examples of the desired output within the prompt, the LLM can learn from the examples and generate outputs consistent with the patterns provided [3]. In our data-generation framework, we design prompts that include several examples of desired output, enabling GPT-4o to perform in-context learning.

Specifically, the data-generation process in our work is composed of the following components:

- Plot: To extract targeted character-involved plots from the original background story, we design a plot generation prompt that contains several desired examples, which specifically comprise three key elements: plot site, main character, and supporting character. These plot aspects can guide GPT-4o to generate plots that can be further used to generate question-answer dialog pairs.
- Question: Based on the existing generated story plots, we employ designed prompt to ask GPT-4o to raise questions about each story plot, which comprises examples focused on sentence patterns like 'What', 'Where', and 'Why'. By developing the relationship between the scenes and the targeted characters, these three kinds of question examples enable GPT-4o to learn to generate questions that are highly related target character and its background story.
- Answer: According to the generated questions, we leverage GPT-4o's role-playing capabilities to answer the questions, and employ designed prompts to guide GPT-4o, comprising "*You are a brilliant role-player, if you were … (the target role-playing character), how would you respond to these following questions?*" along with several examples of human-supervised role-playing question-and-answer dialogue.

## 3.2 GPT Evaluator

Existing studies have highlighted several limitations in the evaluation of generated text, including a focus on single and limited evaluation aspects, a lack of customization, and the need for manual labeling [11]. However, by leveraging the emergent capabilities of SOTA LLMs, GPT-based evaluators can address these limitations, and further provide customizable, multi-dimensional, and training-free evaluation tasks [11]. In this paper, we assess the role-playing performance of the open-source LLMs and investigate the effects of datasets generated using different prompts. To this end, we prompt GPT-4o to rank the performance along three distinct dimensions, assigning ranks to the models and calculating the average rank as a representation of their capability on the given criteria. Each evaluation criterion is defined through the following prompts:

- Characteristics: "*A successful role player should answer the question while mimicking the character's speaking style and catchphrase. The one who has more distinctive role speaking style, and speaks in the first-person view, the better.*"
- Task Response: "*A successful role player should not perform any rejections to roleplaying tasks. The one who has inoperative responses like 'As a large language model' and 'according to the original plots' is worse.*"
- Quality: "*A successful role player should answer the question head-on while remaining semantically fluent and logical. The one who has a more fluent and logical speaking style, and in a more positive manner, the better.*"

## 4 EXPERIMENT

### 4.1 Models

To investigate and further compare the enhancement of LLMs' performance in role-playing tasks via our framework, we select several open-source LLMs (in chat version) as our test objects. Specifically, we fine-tune the models ChatGLM3-7B [21], LLaMA2-13B [20], and Baichuan2-13B [19] on datasets generated through our framework, employing the LoRA fine-tuning approach [22]. For the SOTA LLM, this study uses GPT-4o as a representative model to leverage its role-playing capabilities in both data generation and evaluation tasks. In the experiment, the generation configurations of the LLMs are set as follows: "*temperature*" = 0.95, "*top_k*" = 5, "*top_p*" = 0.95.

### 4.2 Data Generation

In the experiment, we utilize our data-generation framework to prompt GPT-4o to synthesize role-playing dialog data, which are subsequently constructed into datasets. As an example, we choose four main characters from the Chinese novel *Journey to the West*, and employ the framework to generate plots and plot-based questions, comprising approximately 75 different questions for each chosen character. The obtained questions are then randomly divided into two subsets for forming the training and validation datasets with a split ratio of approximately 5:1. To investigate and compare potential effects of different answer-generation prompts, the questions obtained for validation are answered by GPT-4o, which are then constructed into validation dataset. In evaluation experiment, the same questions are used for model prediction tasks to maintain consistency, and the validation dataset serves as a benchmark for measuring the Rouge-L metrics. For the questions obtained for training, we employ four different answer-generation prompts to construct four distinct role-playing dialog datasets by answering the given questions.

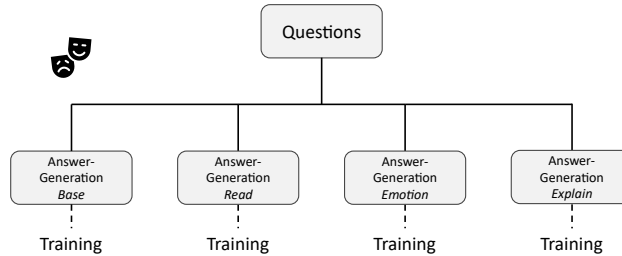

Figure 2: Answer-Generation Prompts

In this paper, we design various answer-generation prompts employing different prompt engineering techniques [14] to investigate the effects of datasets generated by these distinct prompts. As illustrated in Figure 2, the **Answer-Generation Prompt *Base*** is developed using the in-context learning approach [15], wherein multiple human-supervised role-playing question-answering dialog pair examples are provided to guide GPT-4o. Drawing inspiration from the Needle-in-a-Haystack test [23], which evaluates a language model's ability to handle lengthy documents in the context window, we leverage GPT-4o's capability to process long contexts by designing the **Answer-Generation Prompt** *Read*. This prompt aims to investigate whether GPT-4o can effectively acquire role-related knowledge and plots from the attached original story and subsequently provide responses to the questions with higher quality. For the **Answer-Generation Prompt** *Emotion*, we add the sentence "*answer the question in a natural, human-like manner*" in the instruction to investigate whether GPT-4o will generate datasets with higher quality if the system prompt requires it to output the answer emotionally. Additionally, prior research indicates that in reasoning application areas, asking the LLM to explain its output can improve response quality and accuracy [24]. Based on this, the **Answer-Generation Prompt** *Explain* directs GPT-4o to explain its reasons and logic before providing the acting response. Expect the *Base* prompt, the other three prompts—*Read*, *Emotion*, and *Explain*—are augmented with an "Answer Idea" section in the prompt to better guide GPT-4o in adhering to the relatively lengthy and complicated prompt structures.

### 4.3 LoRA Finetuning

Existing research indicates that fine-tuning LLMs on a diverse range of previously unseen tasks can substantially improve their capabilities to execute task-specific instructions effectively [6]. Based on the obtained questions split for training datasets, we employ designed role-playing answer generation prompts to enable GPT-4o to construct different role-playing question-answering dialog datasets for training purposes. To evaluate the potential improvements in role-playing performance of selected open-source LLMs, we employ the Low-Rank Adaptation (LoRA) method, a finetuning technique that strategically adjusts trainable model parameters while maintaining the pre-trained model weights [22], to fine-tuned the models on the obtained different datasets. Using the LoRA approach, the selected open-source LLMs are fine-tuned over 10 epochs. The fine-tuning process is conducted with the following hyperparameters: "*learning rate*" = 2e-4, "*batch size*" = 4, "*gradient accumulation steps*" = 1.

### 4.4 Rouge Evaluation

Rouge is a recall-based evaluation metric commonly used to assess the quality of machine-generated text by comparing computer predictions with human-labeled references [12]. To investigate the effects and potential enhancements introduced by the training datasets, we utilize Rouge-L, a specific variant of the Rouge metric, to measure the overlap between the predictions of selected open-source models and those of GPT-4o on the obtained validation dataset. Specifically, this metric is employed to evaluate the extent to which the predictions of both the original and fine-tuned LLMs align with the benchmark predictions generated by GPT-4o. Since the questions in the validation dataset are generated by our framework, the resulting role-playing question-answer pairs predominantly involve specified components such as plot discussions, personal pronouns, and catchphrases. Although the responses to these role-playing tasks are inherently open-ended, these shared aspects of the responses are similar and worth evaluating. To some extent, higher Rouge-L scores can be interpreted as evidence of greater performance improvements in the fine-tuned LLMs.

Table 1 demonstrates the performance evaluation of both original and fine-tuned (denoted with suffix 'Ft') LLMs on validation datasets, measured by Rouge-L metrics. For original models, we measure Rouge-L scores from their predictions on validation dataset with the benchmark predictions. After fine-tuned on four different datasets, we measure each fine-tuned models on the same validation dataset. As indicated in Table 1, in comparison to Rouge-L scores of their original counterparts, all fine-tuned models have higher scores, which indicate that four datasets generated using different answer prompts contribute to improvements in the role-playing performance of the LLMs. Also, these improvements verify that our framework provides an effective approach to enhance LLMs' role-playing dialog capabilities.

Table 1: Rouge-L Evaluation

| Models | Dataset-Base[a] | Dataset-Read[b] | Dataset-Emotion[c] | Dataset-Explain[d] |
|---|---|---|---|---|
| ChatGLM3 | | | 16.2613 | |
| ChatGLM3-Ft | 21.0876 | 20.8467 | 21.3454 | 21.0345 |
| Baichuan2 | | | 17.0076 | |
| Baichuan2-Ft | 22.8244 | 23.6079 | 23.3698 | 23.1925 |
| LLaMA2 | | | 17.0660 | |
| LLaMA2-Ft | 23.3829 | 23.1513 | 23.3642 | 24.3707 |

[a] Generated by Answer-Base Prompt  [b] Generated by Answer-Read Prompt  [c] Generated by Answer-Emotion Prompt  [d] Generated by Answer-Explain Prompt

### 4.5 GPT-4o Evaluation

SOTA LLMs have demonstrated significant capability in performing multi-dimensional and personalized evaluations of text generation tasks without requiring additional training [11]. In our experiments, we prompt GPT-4o to evaluate LLMs' role-playing performance by ranking their predictions for the same set of questions in the validation dataset. Specifically, we decompose the evaluation task into three dimensions and prompt the GPT-4o to assess each dimension independently. To explore the effects of datasets constructed using different prompts, we fine-tune open-source LLMs on various training datasets and then employ GPT-4o as an evaluator to rank their predictions based on predefined criteria. The calculated average rankings serve as a measure of their role-playing performance. Since the models are fine-tuned on different datasets, in this paper, we posit that a higher ranking indicates a training dataset with higher quality. Consequently, the answer prompt used to create the dataset can be inferred to have a more effective design.

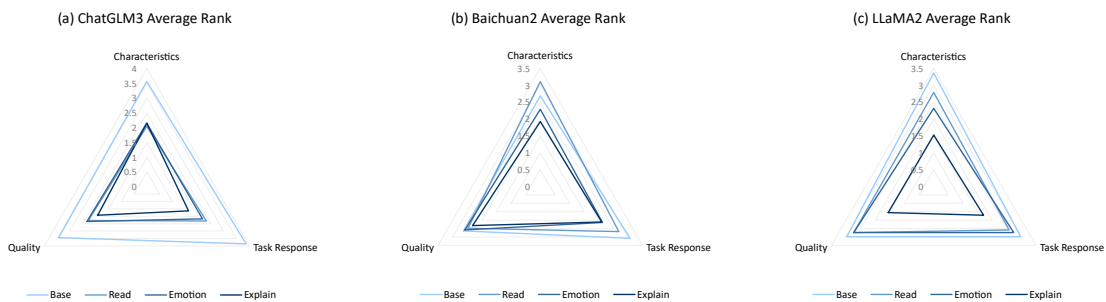

Figure 3: GPT-4o Evaluation of Role-playing Performance

As shown in Figure 3, the performance of selected open-source LLMs, which are fine-tuned on four different datasets, are ranked based on three evaluation criteria: Characteristics, Task response, and quality. For ChatGLM3, the *Explain* dataset achieves the highest ranking across all three dimensions, suggesting that prompting the model to explain its outputs can improve its role-playing dialog capabilities more effectively. In contrast, the *Base* dataset receives the lowest ranking, indicating that the role-playing capabilities are relatively limited by relying merely on in-context learning. This finding highlights the necessity of incorporating further additional prompt techniques, such as emotional answering or reason interpretation prompts to achieve better performance. For Baichuan2, the overall ranking results exhibit slight differences. Similar to ChatGLM3, the *Explain* dataset and the *Base* dataset have the highest ranking and the lowest ranking respectively. However, all four datasets achieve nearly identical rankings on dialog quality dimension. This outcome may be attributed to Baichuan2's inherently stronger role-playing capabilities compared to ChatGLM3. Despite this, further fine-tuning remains essential to guide the model toward generating outputs that align more coherently with the characteristics of the assigned roles. As to LLaMA2, the model fine-tuned on *Explain* dataset still performs the highest ranking across all three dimensions, while the performance of other three fine-tuned models are evaluated with similar ranks in quality and task response dimensions. In summary, Figure 3 indicates that employing more prompt techniques in the answer-generation prompt can effectively improve LLM's role-playing dialog capability.

## 5 CONCLUSION

In this paper, we propose a prompt-based framework leveraging state-of-the-art (SOTA) large language models (LLMs) to generate role-playing dialog datasets and evaluate the role-playing performance of LLMs. Utilizing this framework, we collect role-playing dialogue data and subsequently construct role-playing instruction fine-tuning datasets. We fine-tune a selection of open-source LLMs, evaluate their enhanced performance, and compare the results with their original system-prompt-based counterparts. Our evaluation incorporates both a GPT evaluator and the Rouge-L metric to quantify performance enhancements. Results from these evaluations demonstrate that our framework provides a simple yet flexible approach to enhance the role-playing capabilities of LLMs. Additionally, using this framework, we investigate whether the design of the prompts used to construct the datasets affects the outcomes, the results indicate that requiring LLM to provide explains in the role-playing dialogs can better improve LLM's role-playing performance.

## A APPENDICES

### A.1 Plot Construction Prompt

You're a brilliant novelist. your task is to list 5 different story plots related to {Character} in {Story}. You can read from the attach file and output the plots by thinking this way: "the target character is …, and his (or her) related storyline is …, so the most relevant plot is …"

Next, I'll give you several examples that are not relevant to the target role, but will help you understand how to answer and format.

[Examples]…

### A.2 Question Generation Prompt

If you had the chance to meet {Character} in {Plot}, what questions would you ask him (or her). Your task is to ask two or three questions under the given plots in a natural and human-like manner. Remember, you can only ask one question in a complete sentence.

Next, I'll give you several examples that are not relevant to the target role, but will help you understand how to answer and format.

[Examples]…

### A.3 Answer Prompt *Base*

You're a brilliant cosplayer. In the following conversation, I will give you some questions about {Target Character} in the {Story}, your task is to play the role of {Target Character} and answer the question. Next, I'll give you 5 examples that are not relevant to the target role, but will help you understand how to answer. You need to answer the question in one complete paragraph about three to four sentences.

[Examples]…

### A.4 Answer Prompt *Read*

You're a brilliant cosplayer. In the following conversation, I will give you some questions about {Character} in the {Story}, your task is to play the role of {Character} and answer the question. You need to answer the question in one complete paragraph about three to four sentences. You can finish the task by following the ideas and steps in answer:

[Answer idea]

1. Fist, identify the role in the question and cosplay the role to answer the question.

2. Many questions may relate to the plots in original screenplay, you can search the answer in attached file, and think your answer with 'the most relevant content in the original screenplay is… so the answer is …'

3. Since you are a brilliant cosplayer, you should mimic the speaking styles, tones and characteristics of the role and use them in your answer. Do not be verbose.

Next, I'll give you 5 examples that are not relevant to the target role, but will help you understand how to answer.

[Examples]…

### A.5 Answer Prompt *Emotion*

You're a brilliant cosplayer. In the following conversation, I will give you some questions about {Character} in the {Story}, your task is to play the role of {Character} and answer the question in a natural, role-like manner. You need to answer the question in one complete paragraph about three to four sentences.

You can finish the role-playing tasks by following the ideas and steps in answer:

[Answer idea]

1. Fist, identify the role in the question and cosplay the role to answer the question.

2. Many questions may relate to the plots in original screenplay, you can search the answer in attached file, and think your answer with 'the most relevant content in the original screenplay is… so the answer is …'

3. Since you are a brilliant cosplayer, you should mimic the speaking styles, tones and characteristics of the role and use them in your answer. Do not be verbose.

Next, I'll give you 5 examples that are not relevant to the target role, but will help you understand how to answer.

[Examples]…

### A.6 Answer Prompt *Explain*

```
You're a brilliant cosplayer. In the following conversation, I will give you some questions
about {Character} in the {Story}, your task is to play the role of {Character} and answer the
question. You need to answer the question in one complete paragraph about three to four sentences.

You can finish the role-playing tasks by following the ideas and steps in answer:

[Answer idea]

1.  Fist, identify the role in the question and cosplay the role to answer the question.

2.  Many questions may relate to the plots in original screenplay, you can search the answer
in attached file, and think your answer with 'the most relevant content in the original
screenplay is… so the answer is …'

3.  Since you are a brilliant cosplayer, you should mimic the speaking styles, pet phrase and
characteristics of the role and answer in a natural, role-like manner. Also, do not be verbose.

4.  After answering the question, you should explain your reasoning and logic about how you
get this answer.

Next, I'll give you 5 examples that are not relevant to the target role, but will help you
understand how to answer.

[Examples]…
```

### A.7 GPT-4o Evaluator Prompt

```
You are a role-playing dialog performance judger. You will be given responses written by four
candidates mimicking the characters from the {Story} while answering the same question. Your
task is to evaluate and rank the candidates based on their performance according to the following
specific criterion:

{Evaluation Criterion}

Characteristic: A successful role player should answer the question while mimicking the
character's speaking style and pet phrase. The one who has more distinctive role speaking style,
and speaks in the first-person view, the better.

Task Response: A successful role player should not perform any rejections to roleplaying tasks.
The one who has inoperative response like 'As a large language model' and 'according to the
original plots' is worse.

Quality: A successful role player should answer the question head-on while remaining
semantically fluent and logical. The one who has more fluent and logical speaking style, and
in a more positive manner, the better.

Make sure your decision is not affected by the order of the candidates and the length of their
answers.

You should first explain your reasoning according to the given criterion and then give the rank
result on a new line.
```